\documentclass{article}
\usepackage{spconf,amsmath,graphicx}
\usepackage{array}
\usepackage{nccmath}
\usepackage{color}
\usepackage{cite}
\usepackage{scalerel}
\usepackage{enumitem} % for alphanumerical enumeration

\usepackage{tikz}
\usetikzlibrary{svg.path}
\usepackage{hyperref}
\definecolor{orcidlogocol}{HTML}{A6CE39}
\tikzset{
  orcidlogo/.pic={
    \fill[orcidlogocol] svg{M256,128c0,70.7-57.3,128-128,128C57.3,256,0,198.7,0,128C0,57.3,57.3,0,128,0C198.7,0,256,57.3,256,128z};
    \fill[white] svg{M86.3,186.2H70.9V79.1h15.4v48.4V186.2z}
                 svg{M108.9,79.1h41.6c39.6,0,57,28.3,57,53.6c0,27.5-21.5,53.6-56.8,53.6h-41.8V79.1z M124.3,172.4h24.5c34.9,0,42.9-26.5,42.9-39.7c0-21.5-13.7-39.7-43.7-39.7h-23.7V172.4z}
                 svg{M88.7,56.8c0,5.5-4.5,10.1-10.1,10.1c-5.6,0-10.1-4.6-10.1-10.1c0-5.6,4.5-10.1,10.1-10.1C84.2,46.7,88.7,51.3,88.7,56.8z};
  }
}
\newcommand\orcidicon[1]{\href{https://orcid.org/#1}{\mbox{\scalerel*{
\begin{tikzpicture}[yscale=-1,transform shape]
\pic{orcidlogo};
\end{tikzpicture}
}{|}}}}

\title{PAT++: a cautionary tale about generative visual augmentation for Object Re-identification}
\name{Leonardo Santiago Benitez Pereira\orcidicon{0000-0001-9429-7308}, Arathy Jeevan} 
\address{Universidad Autonoma de Madrid}

\begin{document}
%\ninept
%
\maketitle
\begin{abstract}
Generative data augmentation has demonstrated gains in several vision tasks, but its impact on object re-identification - where preserving fine-grained visual details is essential - remains largely unexplored. In this work, we assess the effectiveness of identity-preserving image generation for object re-identification. Our novel pipeline, named PAT++,  incorporates Diffusion Self-Distillation into the well-established Part-Aware Transformer. Using the Urban Elements ReID Challenge dataset, we conduct extensive experiments with generated images used for both model training and query expansion. Our results show consistent performance degradation, driven by domain shifts and failure to retain identity-defining features. These findings challenge assumptions about the transferability of generative models to fine-grained recognition tasks and expose key limitations in current approaches to visual augmentation for identity-preserving applications.
\end{abstract}
\begin{keywords}
Object Re-Identification, Generative AI, Diffusion Models, Urban Elements Challenge
\end{keywords}
\section{Introduction} \label{sec:intro}

Generative models have recently emerged as powerful tools for data augmentation in computer vision tasks, particularly in scenarios where training datasets are limited or imbalanced. These techniques, which we refer to as generative visual data augmentation \cite{Alimisis2025}, involve leveraging models such as Stable Diffusion to synthesize novel samples. For instance, in a binary classification task distinguishing between platypuses and cats, one may use a text-to-image generative model to generate additional platypus images when the dataset contains a significant class imbalance favoring cats.

However, the application of generative augmentation in identity-centered tasks remains underexplored. In such tasks, the relevant prediction target is not the class of an object, but rather its unique identity \cite{ruiz2023dreambooth}. Examples include object re-identification (ReID) \cite{moral2024urbanelements,he2021transreid}, where the goal is to recognize the same specific object across different images; person-specific data removal \cite{chen2025scoreforgettingdistillation} (e.g., removing all data corresponding to one particular individual such as a specific celebrity); and targeted content generation for specific products \cite{alam2024} (e.g., generating advertising images for one exact purse model). In all these cases, maintaining the visual identity of the object is crucial.

A fundamental challenge arises when applying generative data augmentation to identity-centered applications: augmented images must preserve identity-defining features \cite{Alimisis2025}. In ReID, for instance, introducing artificially generated but identity-inconsistent images can degrade performance, as subtle cues—such as facial details, clothing textures, or surface marks—are essential for differentiating between instances \cite{Zheng2020}. Effective augmentation must therefore maintain these identity-critical features while introducing only permissible variations such as pose, viewpoint, or background.

Several approaches have been proposed to address this challenge. Standard image-conditioned diffusion models, such as ControlNet \cite{zhang2023controlnet}, can guide generation based on a reference image but typically fail to preserve fine-grained identity attributes. DreamBooth \cite{ruiz2023dreambooth} improves fidelity by fine-tuning the generative model on a small set of subject-specific images; however, it is computationally expensive, requiring tens of minutes of finetuning per subject. InstantID \cite{wang2024instantid} introduces a zero-shot alternative by leveraging identity-aware pretraining and facial landmark guidance, but its scope is restricted to human faces. More recently, Diffusion Self-Distillation (DSD) \cite{cai2024dsd} has been proposed as a scalable, general-purpose solution. By pretraining on a dataset of 491,000 paired identity-centered images, DSD enables identity-preserving augmentation across a broad range of categories — including objects, people, and animals — without subject-specific finetuning.

This work investigates the applicability of DSD to the object re-identification task.
The proposed pipeline, named PAT++, builds upon the work of Part-Aware Transformer (PAT) \cite{ni2023} - a reliable and well tested method for ReID tasks - and leverages generated data for both model training (Section \ref{sec_train}) and query expansion (Section \ref{sec_match}). We performed several experiments with the Urban Elements ReID Challenge \cite{urban_reid_challenge}, ablating the effects of utilizing generated data in different stages of the pipeline by comparing the obtained mean Average Precision (mAP) \cite{Zheng2015} in the challenge public evaluation.

The obtained results indicate that even DSD, an State-of-the-Art method for identity preserving image generation, fails to improve the results of PAT. None of the performed experiments lead to a higher mAP than the baseline PAT method, and visual inspection of the generated images indicate that they do not capture important characteristics of the objects, leading to generated images that greatly differ from the original identity.

The contributions of this work are twofold: first, proposing a pipeline that employ generative models for ReID; Second, indicating that identity preserving image generation is still an open research frontier, with even state-of-the-art methods yielding ineffective outcomes. The full implementation is publicly available at Github \footnote{https://github.com/LeonardoSanBenitez/Urban-Elements-ReID-Challenge}.

\section{Literature review}
\subsection{Urban Elements ReID Challenge}
While the re-identification of people and vehicles has received considerable attention \cite{moral2024urbanelements}, the Urban Elements ReID Challenge introduces a novel and underexplored task: long-term re-identification of static urban elements such as trash bins, waste containers, and crosswalks. Unlike previous benchmarks, it provides a unique dataset captured over four months along a consistent urban route, enabling the study of visual consistency and degradation over time \cite{urban_reid_challenge}. 

This long-term perspective introduces temporal challenges such as wear, lighting variation, and occlusions, which are seldom addressed in existing ReID research. By focusing on static infrastructure, this challenge presents new opportunities for applying ReID in smart city applications, such as automated waste management, infrastructure monitoring, and public safety — areas where reliable identification of urban elements is essential.

% TODO: include a small description of the dataset

\subsection{Diffusion Self-Distillation}
DSD uses Flux1.0 \cite{flux2025} as the generative backbone, a 12 billion parameter rectified flow transformer developed by Black Forest Labs. The Flux architecture is adapted to allow image-conditioning - alongside the previously existing text-conditioning - and trained in a supervised manner with triplets conditioning prompt + conditioning image + target noisy image. The training dataset is composed of 491k such triplets, generated automatically with a frozen Flux1.0 (hence the “self-distillation” part of the name) and curated with a combination of GPT-4o-mini (for generating identity-centered prompts) and Gemini-1.5 (for selecting the best paired images).

\subsection{Visual Query Expansion}
The work of Zheng et al. (2020) \cite{Zheng2020} explore visual query expansion in the context of vehicle re-identification by updating the original query feature with the average of its top-ranked neighbors’ features, reinforcing the representation with semantically similar samples. This approach improves retrieval performance by leveraging the structure of the feature space without additional supervision. Notably, the authors carefully exclude low-resolution images from this process to avoid degrading the query representation.

A similar idea is pursued by Gordo et al. (2017) \cite{gordo2017}, who refine query embeddings using feature aggregation in instance-level image retrieval \cite{manning2008}, showing that even modest adjustments to the query vector can significantly improve ranking accuracy. Despite these advances, the design of effective query expansion strategies remains largely empirical and sensitive to dataset-specific characteristics. The absence of principled, generalizable approaches highlights that visual query expansion is still a challenging and evolving area in image retrieval and re-identification.

\section{Methodology}

For each of the query, gallery, and train sets, augmentations (also referred to as refinements) were generated for the following 3 conditions:
\begin{enumerate}[label=\Alph*., itemsep=-1pt]
    \item With a 2 people walking by and partially occluding it;
    \item At sunset with a warm sun shining on it;
    \item On a rainy day, gray sky, low visibility.
\end{enumerate}

To increase the quality of the prompts, each image was classified (that is, identified as trash bin, waste container, or crosswalk) and captioned (that is, a short text describing physical characteristics that make that object unique). To perform such a task, the visual foundational model GPT-4o-mini \cite{openai2024gpt4technicalreport} was utilized
%\cite{azure_ai_foundry}
and prompted with a zero-shot strategy.

Then, a DSD model generates the refinements and saves them to disk. The hyperamaters were empirically adjusted to the values of 3.5 guidance, 1.0 image guidance, and 1.0 text guidance. No post-processing was applied. The generated images were visually inspected and some results were manually selected for discussion and compiled in Table \ref{tab_image_augmentations}. The generation was performed on a Nvidia A100 GPU with 40GB of memory.
The model GPT-4o-mini \cite{openai2024gpt4technicalreport} was deployed using Azure AI Foundry Services.

\begin{table*}[htb]
    \centering
    \renewcommand{\arraystretch}{1.0}
    \setlength{\tabcolsep}{2pt} % adjust as needed

    \begin{tabular}{| >{\centering\arraybackslash}m{0.19\textwidth} 
                    | >{\centering\arraybackslash}m{0.19\textwidth} 
                    | >{\centering\arraybackslash}m{0.19\textwidth} 
                    | >{\centering\arraybackslash}m{0.19\textwidth} 
                    | >{\raggedright\arraybackslash}m{0.24\textwidth}
                    | }

    \hline
    \textbf{Original Image} & \textbf{Augmentation A} & \textbf{Augmentation B} & \textbf{Augmentation C} & \multicolumn{1}{c|}{\textbf{Discussion}} \\
    \hline
    \includegraphics[width=\linewidth,height=\linewidth,keepaspectratio]{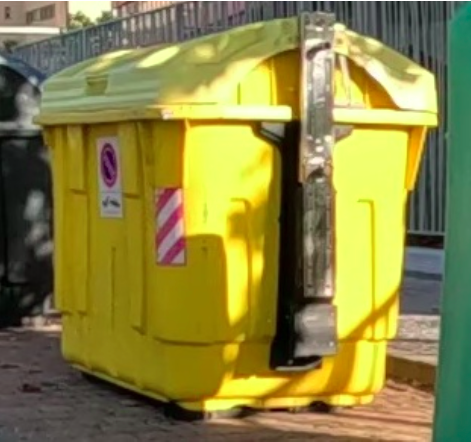} &
    \includegraphics[width=\linewidth,height=\linewidth,keepaspectratio]{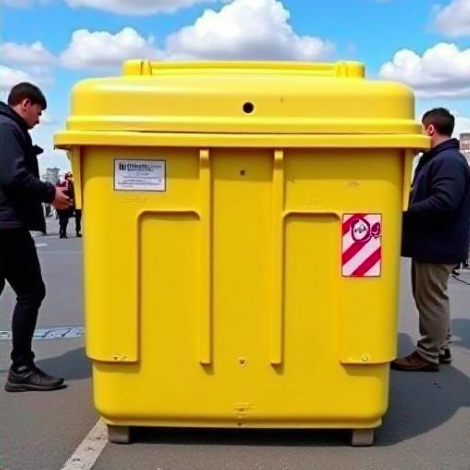} &
    \includegraphics[width=\linewidth,height=\linewidth,keepaspectratio]{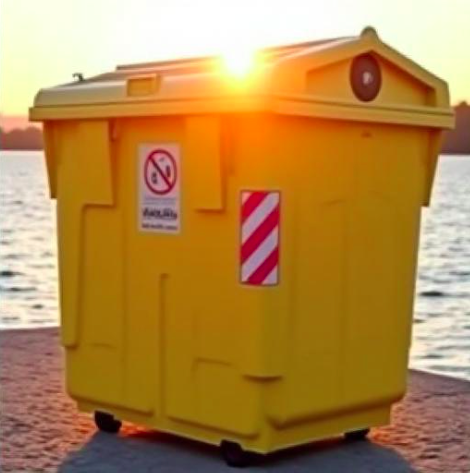} &
    \includegraphics[width=\linewidth,height=\linewidth,keepaspectratio]{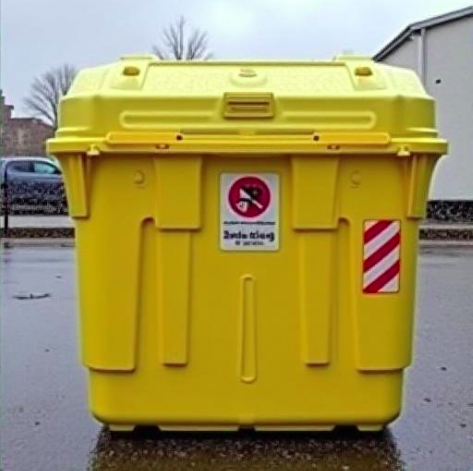} &
    The generated images are similar to the original container, but some important elements are missing (such as the black column on the right side). \\

    \hline
    \includegraphics[width=0.1\linewidth,height=\linewidth,keepaspectratio]{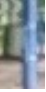} &
    \includegraphics[width=\linewidth,height=\linewidth,keepaspectratio]{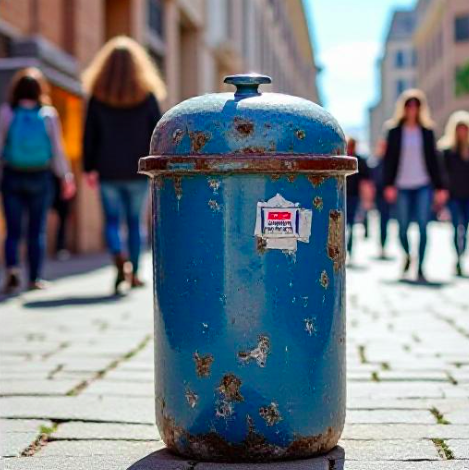} &
    \includegraphics[width=\linewidth,height=\linewidth,keepaspectratio]{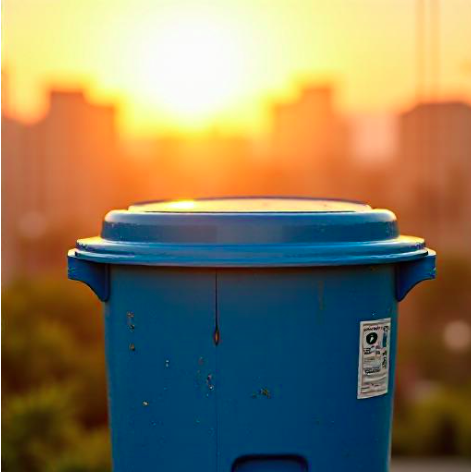} &
    \includegraphics[width=\linewidth,height=\linewidth,keepaspectratio]{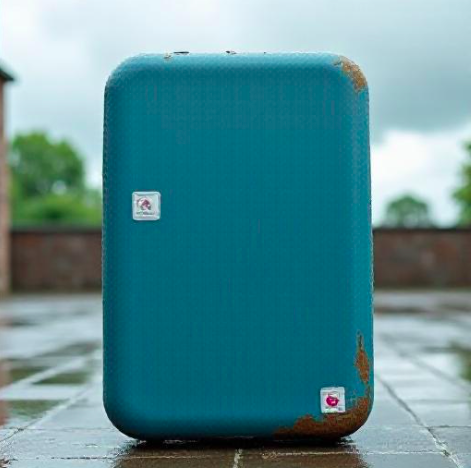} &
    The generated bins are unrelated to the original object, and are highly different among themselves. One possible reason is the extreme low-resolution of the source object. \\

    \hline
    \includegraphics[width=0.5\linewidth,height=\linewidth,keepaspectratio]{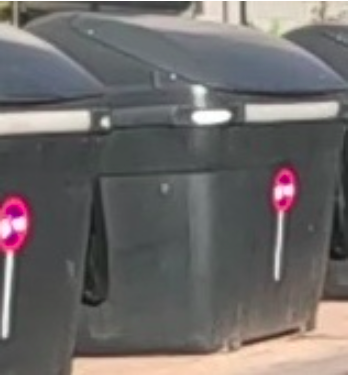} &
    \includegraphics[width=\linewidth,height=\linewidth,keepaspectratio]{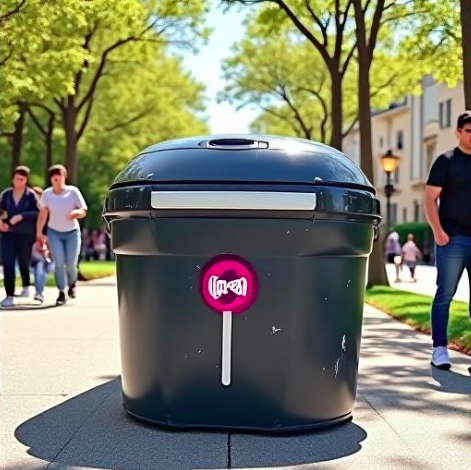} &
    \includegraphics[width=\linewidth,height=\linewidth,keepaspectratio]{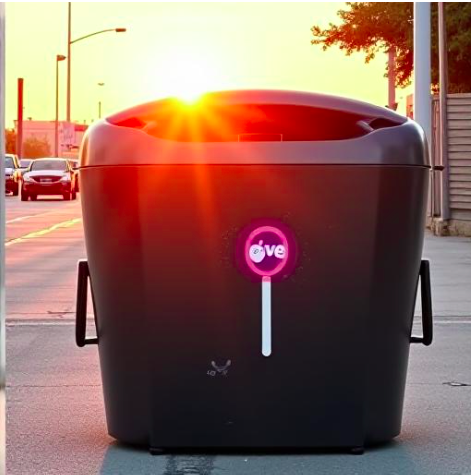} &
    \includegraphics[width=\linewidth,height=\linewidth,keepaspectratio]{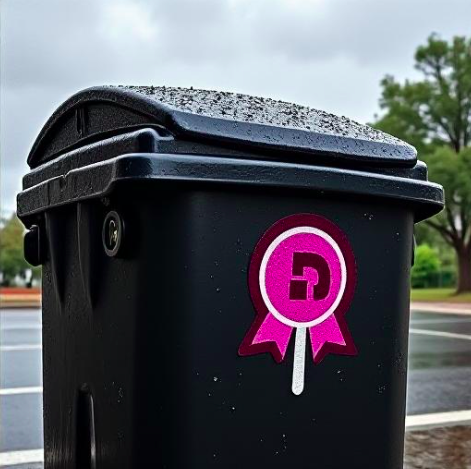} &
    The generated sticker is superficially similar, but clearly not the same. Since many containers have a very similar sticker, the generated images fail to capture the unique characteristics of the source object.\\

    \hline
    \end{tabular}
    \caption{Comparison of augmentations applied to the original image.}
    \label{tab_image_augmentations}
\end{table*}

\subsection{Model training} \label{sec_train}

\begin{figure}[htb]
    \centering
    \includegraphics[width=0.6\linewidth]{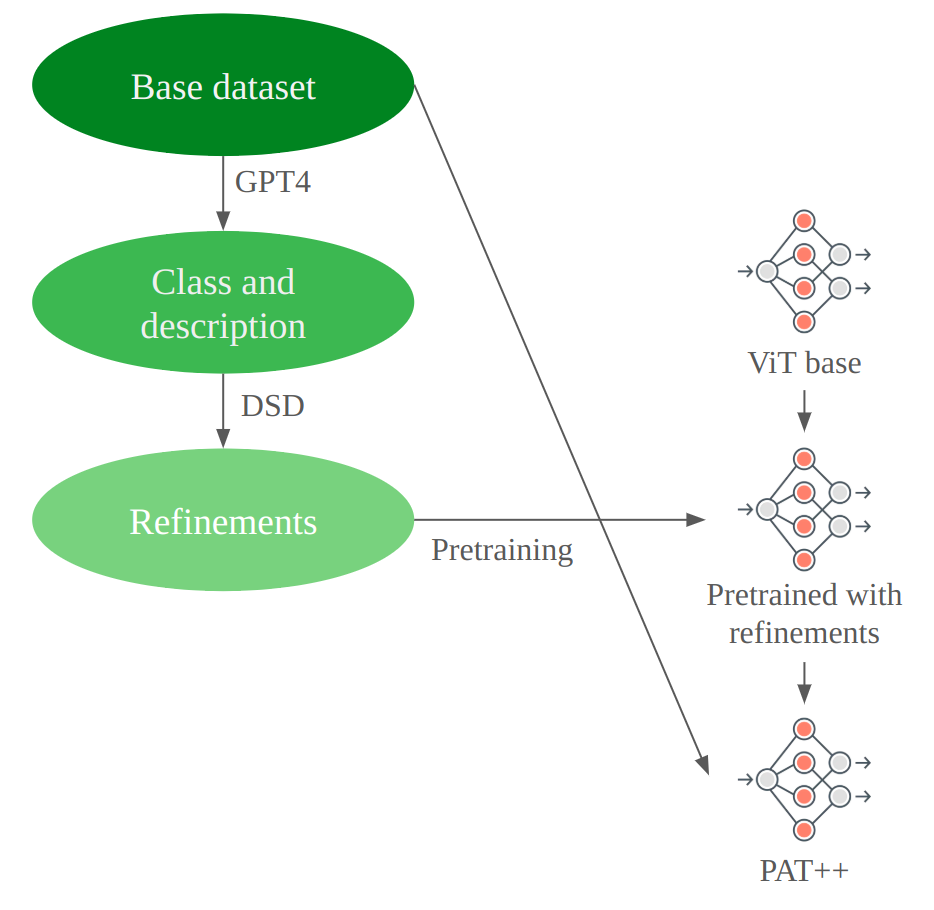}
    \caption{Training process when using first generated images and then real images. Both training stages were performed with the same amount of data and with the same hyperparameters, differing only if they used generated or real data.}
    \label{fig_method_train}
\end{figure}

Training the PAT++ embedding model was performed in two stages: first, starting with a pre-trained ViT-B model and finetuning using only the generated data; Second, continuing the finetuning using only real images. Such separation was chosen because the generated images have a different style (smooth surfaces, ideal light conditions) and quality (high-resolution, well-focused, no noise) compared to the original images, and such domain shift could harm the learning. The process is illustrated in Figure \ref{fig_method_train}.

Even though the first step could use an arbitrary amount of generated data, we choose to use the same amount as the available real data. In both steps, the training was performed for 60 epochs with 0.001 learning rate and Stochastic Gradient Descent optimizer. The official repository for Part-Aware-Transformer \footnote{https://github.com/liyuke65535/Part-Aware-Transformer} was used, with the provided pre-trained model ViT-B. The official repository for DSD \footnote{https://github.com/primecai/diffusion-self-distillation} was utilized, also with the provided pre-trained model.

\subsection{Query expansion} \label{sec_match}

Leveraging the augmented images for improving the matching and post-processing of the ReID pipeline can be done independently from using PAT++ (that is, with or without training the embedding model on generated images), since all that is needed is to calculate the feature vector of the refined images with any embedding model. For simplicity, this Section describes the pipeline by referring to the embedding model as PAT++.

\begin{figure}[htb]
    \centering
    \includegraphics[width=1\linewidth]{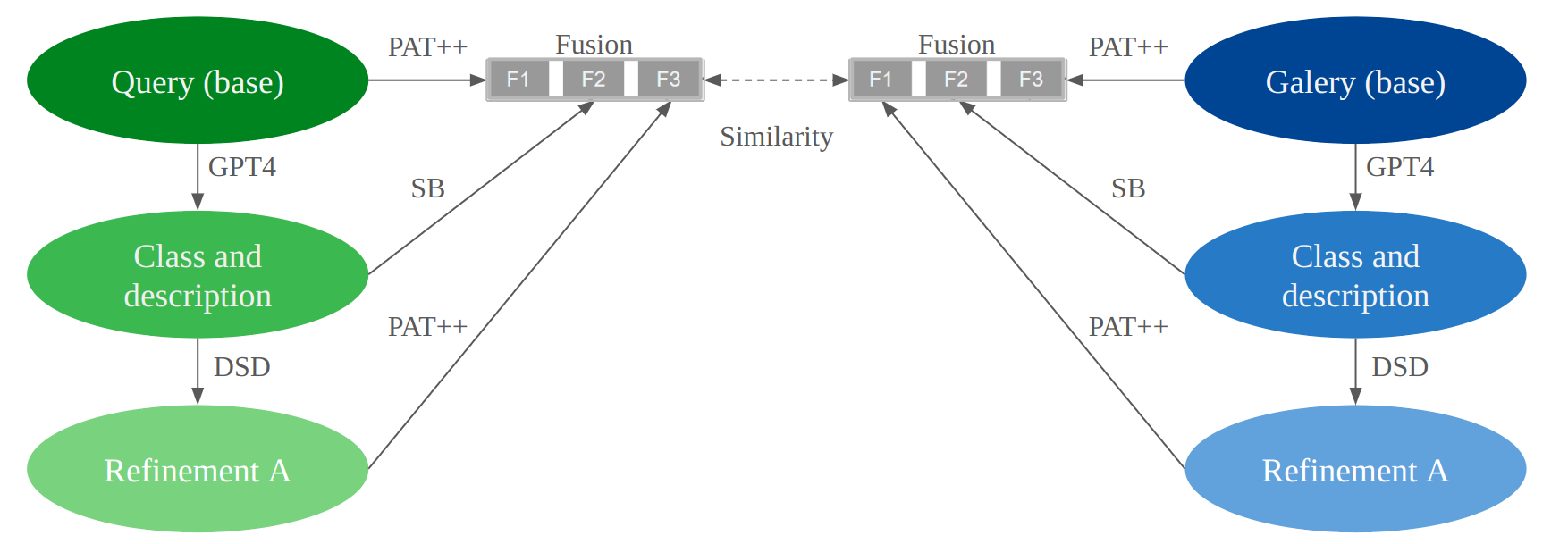}
    \caption{Matching process when utilizing PAT++ embeddings for both the base (original) and refined (generated) images, as well as SB for the textual class and description. Only one generated image (Refinement A) is shown in the diagram, but multiple can be used.}
    \label{fig_method_match}
\end{figure}

The image classification and description obtained with GPT-4o-mini can optionally be utilized for the feature vector, by using a text embedding model such as Sentence Bert (SB) \cite{reimers2019sentence_bert}. To ease the pipeline description, this Section assumes SB is used to vectorize the class and description. These embedding vectors - for the original images, for the text, and for the refined images - can be fused in different ways, such as averaging or concatenating them. After fusion, the pipeline is exactly as described in depth in other works \cite{ni2023}, and summarized in Figure \ref{fig_method_match}.

\section{Experiments}

Since there are many design choices that affect the performance of the ReID pipeline, we systematically tested several combinations of these choices and reported the most interesting results in Table \ref{table_comparison}. For all experiments, the best hyperparameters (for the training as well as for the query expansion) were chosen empirically after several tests. For all experiments, cosine similarity was used to match the query and gallery vectors, and k-reciprocal encoding was used to rerank the final results \cite{zhong2017rerank_k_reciprocal}.
With respect to the method used to calculate the embedding features of the base and refined images, the following methods were tested:
\begin{itemize}[itemsep=-5pt]
    \item One single PAT model (trained only with real images) for both base and refinement;
    \item Separate PAT models for real and generated images, trained respectively only with only real and generated images;
    \item Single PAT++ embedding for all images, trained as described in Section \ref{sec_train};
    \item Using Dino V2 Base \cite{oquab2024dinov2} for embedding the generated images;
    \item Using 1 or 3 query refinements;
    \item With and without Sentence Bert for embedding the textual class and description.
\end{itemize}

Furthermore, these different embeddings must be fused together before performing the matching between query and gallery images. The tested fusion methods were:
\begin{itemize}[itemsep=-5pt]
    \item Average, both simple and weighted;
    \item Concatenate, both simple and weighted;
    \item Use refinements only when base similarity is below its 20th percentile;
    \item Match just refinement-with-refinement and base-with-base (no fusion).
\end{itemize}

\begin{table}[htb]
    \centering
    \begin{tabular}{|l|l|l|l|}
        \hline
        \textbf{E. Base} & \textbf{E. Refinements} & \textbf{Fusion} & \textbf{mPA} \\ \hline
        PAT & None    & None        & -0.0\%           \\ \hline
        PAT & PAT     & Average     & -2.923\%     \\ \hline
        PAT & PAT     & Concatenate & -3.123\%     \\ \hline   
        PAT & PAT, SB & Concatenate & -3.245\%     \\ \hline
        PAT & Dino    & Concatenate & -5.115\%     \\ \hline
        PAT & PAT++   & Average     & -3.986\%      \\ \hline
        PAT++ & None  & None        & -3.568\%      \\ \hline
        PAT++ & PAT++ & Average     & -3.405\%      \\ \hline
    \end{tabular}
    \caption{Comparison of the main experiments performed on the Urban Elements ReID Challenge \cite{urban_reid_challenge}, displaying the embedding model used for base (real) images, embedding model used for refined (generated) images, method used for using the embeddings, and mPA relative to baseline (using only PAT, with only real images).}
    \label{table_comparison}
\end{table}

%\pagebreak
\section{Discussion}

These preliminary results indicate that even DSD, the current state-of-the-art in identity-preserving image generation, fails to capture important characteristics of the objects, leading to generated images that greatly differ from the original identity. Another limitation is that the official DSD implementation does not allow for controlling the resolution of the generated image, which would need to be handled in a post-processing step, since - for ReID purposes - it would be ideal if the generated images had the same resolution as the original images, easing the matching.

Another identified problem is that the augmented images look clearly AI-generated, with an unrealistic level of detail and highly exaggerated light conditions. Taken together, these differences can be considered a form of domain shift that harms both the training and the query expansion. Even though there is extensive literature for handling domain shifts, such as using Domain-Invariant Feature Learning \cite{Ganin2016}, exploring such techniques is beyond the scope of this work.

These problems led to every experiment under-performing the baseline. That is, no usage of DSD-generated images improved the results obtained in the Urban Elements ReID Challenge \cite{urban_reid_challenge}.
We can conclude from these findings that utilizing generative visual data augmentation in identity-centered applications is still an open research frontier, with even state-of-the-art methods yielding ineffective outcomes.

To enhance the application of generative augmentation in ReID, future research could investigate post-processing techniques that address discrepancies between synthetic and real images. One approach could involve integrating domain-adaptation techniques — like adversarial feature alignment or style transfer — to reduce domain shift from generated samples and enable fine-grained control during synthesis (such as keypoint-conditioned diffusion) to preserve instance details. Additionally, applying DSD selectively to only high-quality source images, while excluding low-resolution or noisy inputs, could help avoid spurious hallucinations that compromise performance, thus improving the balance between visual augmentation and identity fidelity.

%If the last page of your paper is only partially filled, arrange the columns so that they are evenly balanced if possible, rather than having one long column.

%In LaTeX, to start a new column (but not a new page) and help balance the last-page column lengths, you can use the command ``$\backslash$pagebreak''

% Below is an example of how to insert images. Delete the ``\vspace'' line,
% uncomment the preceding line ``\centerline...'' and replace ``imageX.ps''
% with a suitable PostScript file name.
% -------------------------------------------------------------------------

% To start a new column (but not a new page) and help balance the last-page
% column length use \vfill\pagebreak.
% -------------------------------------------------------------------------
%\vfill
%\pagebreak

\vfill
\pagebreak
\clearpage

% References should be produced using the bibtex program from suitable
% BiBTeX files (here: strings, refs, manuals). The IEEEbib.bst bibliography
% style file from IEEE produces unsorted bibliography list.
% -------------------------------------------------------------------------
\bibliographystyle{IEEEbib}

\begin{thebibliography}{10}

\bibitem{Alimisis2025}
Panagiotis Alimisis, Ioannis Mademlis, Panagiotis Radoglou-Grammatikis, Panagiotis Sarigiannidis, and Georgios~Th. Papadopoulos,
\newblock ``Advances in diffusion models for image data augmentation: a review of methods, models, evaluation metrics and future research directions,''
\newblock {\em Artificial Intelligence Review}, vol. 58, no. 4, pp. 112, jan 2025.

\bibitem{ruiz2023dreambooth}
Nataniel Ruiz, Yuanzhen Li, Varun Jampani, Yael Pritch, Michael Rubinstein, and Kfir Aberman,
\newblock ``Dreambooth: Fine tuning text-to-image diffusion models for subject-driven generation,'' 2023.

\bibitem{moral2024urbanelements}
Paula Moral, Álvaro García-Martín, and José~M. Martínez,
\newblock ``Long-term geo-positioned re-identification dataset of urban elements,''
\newblock in {\em 2024 IEEE International Conference on Image Processing (ICIP)}, 2024, pp. 124--130.

\bibitem{he2021transreid}
Shuting He, Hao Luo, Pichao Wang, Fan Wang, Hao Li, and Wei Jiang,
\newblock ``Transreid: Transformer-based object re-identification,''
\newblock in {\em Proceedings of the IEEE/CVF International Conference on Computer Vision (ICCV)}, October 2021, pp. 15013--15022.

\bibitem{chen2025scoreforgettingdistillation}
Tianqi Chen, Shujian Zhang, and Mingyuan Zhou,
\newblock ``Score forgetting distillation: A swift, data-free method for machine unlearning in disffusion models,'' 2025.

\bibitem{alam2024}
Mohammad~Mahmudul Alam, Negin Sokhandan, and Emmett Goodman,
\newblock ``Automated virtual product placement and assessment in images using diffusion models,'' 2024.

\bibitem{Zheng2020}
Zhedong Zheng, Minyue Jiang, Zhigang Wang, Jian Wang, Zechen Bai, Xuanmeng Zhang, Xin Yu, Xiao Tan, Yi~Yang, Shilei Wen, and Errui Ding,
\newblock ``Going beyond real data: A robust visual representation for vehicle re-identification,''
\newblock in {\em 2020 IEEE/CVF Conference on Computer Vision and Pattern Recognition Workshops (CVPRW)}, June 2020, pp. 2550--2558.

\bibitem{zhang2023controlnet}
Lvmin Zhang, Anyi Rao, and Maneesh Agrawala,
\newblock ``Adding conditional control to text-to-image diffusion models,'' 2023.

\bibitem{wang2024instantid}
Qixun Wang, Xu~Bai, Haofan Wang, Zekui Qin, Anthony Chen, Huaxia Li, Xu~Tang, and Yao Hu,
\newblock ``Instantid: Zero-shot identity-preserving generation in seconds,'' 2024.

\bibitem{cai2024dsd}
Shengqu Cai, Eric Chan, Yunzhi Zhang, Leonidas Guibas, Jiajun Wu, and Gordon. Wetzstein,
\newblock ``Diffusion self-distillation for zero-shot customized image generation,''
\newblock in {\em CVPR}, 2025.

\bibitem{ni2023}
Hao Ni, Yuke Li, Lianli Gao, Heng~Tao Shen, and Jingkuan Song,
\newblock ``Part-aware transformer for generalizable person re-identification,''
\newblock in {\em Proceedings of the IEEE/CVF International Conference on Computer Vision}, 2023, pp. 11280--11289.

\bibitem{urban_reid_challenge}
VPULab,
\newblock ``Urban elements reid challenge,'' \url{https://kaggle.com/competitions/urban-reid-challenge}, 2025,
\newblock Kaggle.

\bibitem{Zheng2015}
Liang Zheng, Liyue Shen, Lu~Tian, Shengjin Wang, Jingdong Wang, and Qi~Tian,
\newblock ``Scalable person re-identification: A benchmark,''
\newblock in {\em 2015 IEEE International Conference on Computer Vision (ICCV)}, Dec 2015, pp. 1116--1124.

\bibitem{flux2025}
{Black Forest Labs},
\newblock ``Flux.1: A new standard in text-to-image generation,'' \url{https://bfl.ai/announcements/24-08-01-bfl}, 2024,
\newblock Accessed: 2025-05-22.

\bibitem{gordo2017}
Albert Gordo, Jon Almazan, Jerome Revaud, and Diane Larlus,
\newblock ``End-to-end learning of deep visual representations for image retrieval,'' 2017.

\bibitem{manning2008}
Christopher~D. Manning, Prabhakar Raghavan, and Hinrich Sch{\"u}tze,
\newblock {\em Introduction to Information Retrieval},
\newblock Cambridge University Press, Cambridge, UK, 2008.

\bibitem{openai2024gpt4technicalreport}
OpenAI,
\newblock ``Gpt-4 technical report,'' 2024.

\bibitem{reimers2019sentence_bert}
Nils Reimers and Iryna Gurevych,
\newblock ``Sentence-bert: Sentence embeddings using siamese bert-networks,''
\newblock in {\em Proceedings of the 2019 Conference on Empirical Methods in Natural Language Processing}. 11 2019, Association for Computational Linguistics.

\bibitem{zhong2017rerank_k_reciprocal}
Zhun Zhong, Liang Zheng, Donglin Cao, and Shaozi Li,
\newblock ``Re-ranking person re-identification with k-reciprocal encoding,'' 2017.

\bibitem{oquab2024dinov2}
Maxime Oquab, Timothée Darcet, Théo Moutakanni, Huy Vo, Marc Szafraniec, Vasil Khalidov, Pierre Fernandez, Daniel Haziza, Francisco Massa, Alaaeldin El-Nouby, Mahmoud Assran, Nicolas Ballas, Wojciech Galuba, Russell Howes, Po-Yao Huang, Shang-Wen Li, Ishan Misra, Michael Rabbat, Vasu Sharma, Gabriel Synnaeve, Hu~Xu, Hervé Jegou, Julien Mairal, Patrick Labatut, Armand Joulin, and Piotr Bojanowski,
\newblock ``Dinov2: Learning robust visual features without supervision,'' 2024.

\bibitem{Ganin2016}
Yaroslav Ganin, Evgeniya Ustinova, Hana Ajakan, Pascal Germain, Hugo Larochelle, Fran\c{c}ois Laviolette, Mario Marchand, and Victor Lempitsky,
\newblock ``Domain-adversarial training of neural networks,''
\newblock {\em J. Mach. Learn. Res.}, vol. 17, no. 1, pp. 2096–2030, Jan. 2016.

\end{thebibliography}

\end{document}